\documentclass[applsci,article,accept,pdftex,moreauthors]{Definitions/mdpi} 
\usepackage{latexsym}
\usepackage{amsmath}
\usepackage{algorithm}
\usepackage{algorithmic}
\usepackage{graphicx}
\usepackage{multirow}
\usepackage{amssymb}
\usepackage{inconsolata}
\usepackage{CJKutf8}
\usepackage{stfloats}
\newcommand{\BIBentryALTinterwordspacing}{\hskip -0.15em plus 0.15em}
\newcommand{\BIBentrySTDinterwordspacing}{\spaceskip=2pt plus 0.5pt minus 0.4pt}
\usepackage{changes} 
\usepackage{arydshln}
\newcommand{\BIBforeignlanguage}[2]{%
  \expandafter\ifx\csname l@#1\endcsname\relax
    \typeout{** WARNING: Unknown language '#1' **}%
    {\it #2}%
  \else
    \foreignlanguage{#1}{#2}%
  \fi
}
\firstpage{1} 
\pubvolume{13}
\issuenum{8}
\articlenumber{4758}
\pubyear{2023}
\copyrightyear{2023}
\externaleditor{Academic Editors: Ahmed Rafea and Julian Szymanski}
\datereceived{10 March 2023} 
\daterevised{3 April 2023}
\dateaccepted{7 April 2023} 
\datepublished{10 April 2023} 
\hreflink{https://doi.org/\\10.3390/app13084758} 

\Title{EvoText: Enhancing Natural Language Generation Models via Self-Escalation Learning for Up-to-Date Knowledge and Improved Performance}

\TitleCitation{EvoText: Enhancing Natural Language Generation Models via Self-Escalation Learning for Up-to-Date Knowledge and Improved Performance}

\Author{{Zhengqing Yuan} 
 $^{1,}$*\orcidA{}, Huiwen Xue $^{2,\dagger}$,  Chao Zhang $^{1,\dagger}$ and Yongming Liu $^{1,}$*$^{,\dagger}$\orcidB{} }

\AuthorNames{Zhengqing Yuan, Huiwen Xue, Chao Zhang and Yongming Liu}

\AuthorCitation{Yuan, Z.; Xue, H.; Zhang, C.; Liu, Y.}

\address{%
$^{1}$ \quad School of Artificial Intelligence, Anhui Polytechnic University, Wuhu 241009, China;\\
$^{2}$ \quad School of Optoelectronic Science and Engineering, Soochow University, Suzhou 215031, China;}

\corres{Correspondence: zhengqingyuan@ieee.org (Z.Y.); liuyongming1015@163.com (Y.L.)}
\firstnote{These authors contributed equally to this work.} 

\abstract{In recent years, pretrained models have been widely used in various fields, including natural language understanding, computer vision, and natural language generation. However, the performance of these language generation models is highly dependent on the model size and the dataset size. While larger models excel in some aspects, they cannot learn up-to-date knowledge and are relatively difficult to relearn. In this paper, we introduce EvoText, a novel training method that enhances the performance of any natural language generation model without requiring additional datasets during the entire training process (although a prior dataset is necessary for pretraining). EvoText employs two models: $G$, a text generation model, and $D$, a model that can determine whether the data generated by $G$ is legitimate. Initially, the fine-tuned $D$ model serves as the knowledge base. The text generated by $G$ is then input to $D$ to determine whether it is legitimate. Finally, $G$ is fine-tuned based on $D$'s output. EvoText enables the model to learn up-to-date knowledge through a self-escalation process that builds on a priori knowledge. When EvoText needs to learn something new, it simply fine-tunes the $D$ model. Our approach applies to autoregressive language modeling for all Transformer classes. With EvoText, eight models achieved stable improvements in seven natural language processing tasks without any changes to the model structure.}

\keyword{training system; fine-tuning; BERT; GPT} 

\begin{document}


\section{Introduction}
\label{sec:introduction}

Pre-training models have shown great promise in natural language processing, with the Transformer model \cite{NIPS2017_3f5ee243} proposing an encoder--decoder architecture based solely on the self-attention mechanism, enabling the construction of large-scale models that can be pretrained on vast amounts of data. Language models~\cite{5393057,9721159,9755057} can be broadly categorized into two types: autoregressive language modeling and autoencoder language modeling. autoregressive language models, such as ELMO~\cite{peters-etal-2018-deep}, GPT~\cite{Radford2018ImprovingLU}, and T5~\cite{2020t5}, predict the next possible word based on the preceding context, making them well-suited for generative tasks. On the other hand, autoencoder language models, such as BERT~\cite{devlinetal2019bert} and RoBERTa~\cite{zhuangetal2021robustly}, predict intermediate words based on context and are better suited for natural language understanding tasks.

In recent years, generative models, including VAE~\cite{kipf2016variational}, GAN~\cite{GANM}, and DDPM~\cite{NEURIPS2020_4c5bcfec}, have made significant progress in computer vision. However, natural language generation presents unique challenges due to its discrete and temporal nature. To address these challenges, a common approach is to use unsupervised training of large language models, such as GPT-3~\cite{brown2020language}, which has 175 billion parameters. Despite their potential, training these models can be challenging due to their size, and further training once deployed can be difficult. As a result, a zero-shot approach is often adopted, which does not require fine-tuning the model for specific downstream tasks. However, this approach has limitations; large language models may not perform as well as smaller models with fine-tuning for certain tasks that rely heavily on supervised learning. It is also akin to using a computer that has not been upgraded for an extended period, and its obsolescence is only a matter of time. Therefore, there is an urgent need for novel approaches that can balance the benefits and limitations of large language models to improve their performance and longevity. While reinforcement learning from human feedback (RLHF)~\cite{christiano2017deep} is one such approach that holds promise, it is still limited by the availability and quality of human feedback data, and it may not always generalize well to other tasks.

\begin{figure}[H]
\includegraphics[width=12cm]{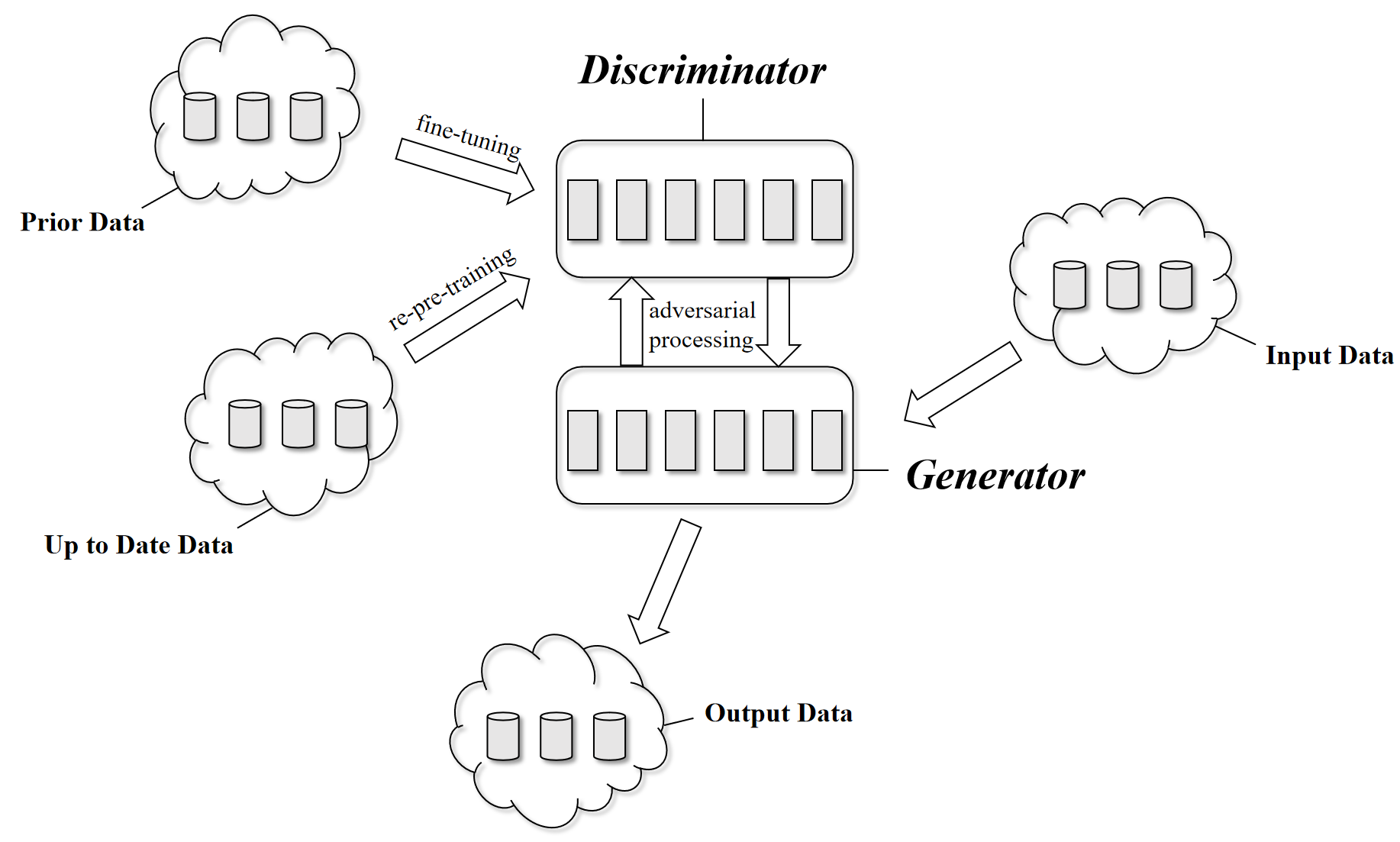}
\caption{EvoText full process.\label{RPN_example1}}
\end{figure}

This paper introduces a novel training process for pretrained models, which we call {\textbf{EvoText}}
. The proposed EvoText method can continuously learn from new data without suffering from the limitations of unsupervised learning or requiring additional datasets for fine-tuning. Specifically, we merge the input text and the generated text and then use a natural language understanding model to label the data for supervised fine-tuning of the generative model. Simply retraining the smaller discriminator model is required to enable the generative model to acquire up-to-date knowledge. To address the issues of natural language understanding errors and overfitting, we adopt a small learning rate and epoch size during fine-tuning. Unlike GAN models, we do not modify the parameters of the natural language understanding model during training, and only fine-tune it when necessary to incorporate new knowledge. The contributions of this work are as follows:
\begin{itemize}

\item EvoText partially mitigates the problem of low-quality samples generated by the generative model.

\item EvoText improves the model's performance without additional data during the training process. (Note that while additional datasets are used in the system for warm-up and learning up-to-date data, as illustrated in Figure \ref{RPN_example1}, only the data generated by the generator are used in the crucial training process.)

\item EvoText enables continuous and sustainable improvement in seven natural language processing tasks, including natural language understanding and natural language generation, without altering the model's structure.

\item The proposed method achieves results comparable to those of large generative networks, even with relatively limited computational resources.

\item We will make the source code for EvoText publicly available on {GitHub} 
 {(\url{https://github.com/DLYuanGod/Auto-learning} (accessed on 9 April 2023).}

\end{itemize}

The novelty of this paper lies in the proposed method for enhancing the performance of natural language generation models without the need for additional datasets during training. With this approach, EvoText allows the generative model to acquire up-to-date knowledge with minimal additional cost. This is achieved by updating the knowledge base through retraining the smaller discriminator model, as opposed to retraining the entire model with additional data.
\section{Background}

In this section, we briefly overview the language model, Transformer, the GPT series of models, the BERT series of models, and the GAN model.

\subsection{Language Modeling}
\textls[-15]{In this subsection, we briefly describe the implementation of the two language modelings.}

\subsubsection{Autoregressive Language Modeling}
Autoregressive language modeling~\cite{6353548,7063240} is a type of language modeling used in natural language processing. It involves predicting the next token in a sequence based on the previous tokens. Given a sequence of N tokens, $\left ( t_{1},t_{2},\dots,t_{N}\right )$, the probability of a token $t_{k}$ is modeled by calculating the conditional probability of $t_{k}$ given all preceding tokens, $\left ( t_{1},t_{2},\dots,t_{k-1}\right)$, using the following formula:
\begin{equation}\label{eq1}
P\left ( t_{1},t_{2},\dots,t_{N} \right ) = \prod_{k=1}^{N} P\left (t_{k} \mid t_{1},t_{2},\dots,t_{k-1} \right )
\end{equation}

This Formula~(\ref{eq1}) calculates the joint probability of all the tokens in the sequence. The product of all the conditional probabilities is taken from the first token to the last token. This means that to predict the probability of a token, we need to know the probability of all preceding tokens. In the case of the backward model, the token after $t_{k}$ needs to be~calculated.

\subsubsection{Autoencoder Language Modeling}

Autoencoder language modeling (ALM)~\cite{7163326,9878160} is a type of language modeling that involves predicting a target token in a sequence of tokens, while considering the probabilities of all preceding and following tokens. This can be represented mathematically as predicting $t_k$ in the sequence $(t_1, t_2, ..., t_k, ..., t_N)$, where the probabilities of $t_{1:k-1}$ and $t_{k+1:N}$ are also~calculated.

ALM is a powerful modeling method that can be applied to natural language understanding tasks. By considering the entire sentence when predicting a target token, ALM can capture the semantic and syntactic relationships between words in the sentence. This makes it particularly useful for tasks such as language generation, summarization, and machine translation.

The concept of ALM is closely related to the masked language model (MLM), which is used as a pretraining strategy in the popular BERT model. In MLM, a percentage of the tokens in a sequence are randomly masked out, and the model is trained to predict the missing tokens while considering the context of the surrounding tokens. This approach is similar to ALM, as it also involves predicting a target token while considering the context of the surrounding tokens.

\subsection{Transformer}
The Transformer model~\cite{9303437,9962811,9585521,9721159,9376902,9733279} is modeled and applied to natural language processing tasks using only self-attentive mechanisms. 

\subsubsection{{Transformer Encoder} 
} Transformer Encoder takes a sequence of tokens as input, which are first processed through a word embedding and positional embedding layer. The resulting vector dimension is called $d_{model}$.

Next, the Transformer Encoder uses a self-attentive mechanism to compute the output tokens. This mechanism involves creating three copies of the input token, which are referred to as \emph{Q}, \emph{K}, and \emph{V}. Each of these copies is used in the attention calculation, which computes the weights between all pairs of tokens in the input sequence. The attention calculation formula is shown below:
\begin{eqnarray}
Attention\left ( Q, K, V \right ) = softmax\left (\frac{QK^{T}}{\sqrt{d_{k}} } \right ) V \nonumber
\end{eqnarray}

Here, $d_{k}$ represents the vector dimension processed by each attention head. The value of $d_{k}$ is equal to $d_{model}$ divided by the number of attention heads used in the multi-head attention mechanism.

Finally, the values calculated for each attention head are concatenated and passed through a multi-layer perceptron (MLP) layer to produce the final output tokens. Overall, the self-attentive mechanism used in the Transformer Encoder allows the model to capture complex relationships between tokens in the input sequence and produce high-quality representations for natural language processing tasks.

\subsubsection{Transformer Decoder} 
The decoder module is calculated in the same way as the encoder module, except that masking is added to the multi-headed attention mechanism to mask tokens that have not yet been generated.

\subsection{GPT Series of Models}

\subsubsection{{GPT}} 
The Generative Pre-training Transformer (GPT)~\cite{Radford2018ImprovingLU} was introduced by Radford et al. in 2018 as an improvement on the Transformer model, which had been mainly used for natural language understanding tasks. GPT was the first model to apply a pretrained Transformer model to natural language processing.

GPT uses a multi-layer Transformer Decoder for the language model, which consists of 12 blocks of Transformer Decoders~\cite{j.2018generating} with up to 117 million parameters. This approach allows GPT to generate high-quality natural language text by predicting the next word in a sequence of words.

Overall, GPT has been shown to achieve impressive results on a range of natural language processing tasks, such as text classification, language translation, and text generation. Its success has led to the development of larger and more powerful Transformer-based models, such as GPT-2 and GPT-3, which continue to push the boundaries of natural language processing.

\subsubsection{{GPT-2 and 
			GPT-3}} GPT-2~\cite{radford2019language} and GPT-3~\cite{brown2020language} represent advanced versions of the original Generative Pre-training Transformer (GPT) model. They employ 48 and 96-layer Transformer Decoder stacks, respectively, with a significantly larger number of parameters: 1.5 billion and 175~billion.

In addition to their larger model sizes, both GPT-2 and GPT-3 are trained on larger and more diverse datasets, enabling them to capture more complex and nuanced patterns in natural language. As a result, these models have achieved state-of-the-art performance on a range of natural language processing tasks, including language translation, question answering, and text generation.

The success of GPT-2 and GPT-3 highlights the tremendous potential of pretrained Transformer models in natural language processing research. With continued advances in this field, we can expect even more powerful language models in the future.

\subsection{BERT Series of Models}

BERT~\cite{devlinetal2019bert,9437636}, RoBERTa~\cite{zhuangetal2021robustly}, ALBERT~\cite{Lan2020ALBERT}, XLNET~\cite{10.5555/3454287.3454804}, TinyBERT~\cite{jiaoetal2020tinybert}, and ELECTRA~\cite{Clark2020ELECTRA} are all state-of-the-art natural language understanding models that employ the Transformer Encoder layer. While there are slight differences in their training approaches, they all share the same underlying architecture.

The architecture of the Transformer Encoder layer is particularly well-suited for natural language understanding tasks, as it allows the model to capture long-range dependencies between words and phrases in the text. By leveraging self-attention mechanisms, these models can dynamically weigh the importance of different words in the input sequence, enabling them to extract more nuanced and complex representations of language.

\subsection{GAN}

The main idea of generative adversarial networks (GAN)~\cite{GANM,10555531570963157346,Isola2016ImagetoImageTW} is to build two models, a generator ($G$) model and a discriminator ($D$) model. During training, the $G$ model tries to improve its manufacturing process to create realistic outputs that can fool the $D$ model, while the $D$ model tries to accurately distinguish between real and generated data like a police officer inspecting forgeries. However, achieving a balance between the two models is essential, since an imbalance can lead to one model not converging. Specifically, the $G$ model needs to create realistic outputs that can fool the $D$ model, while the $D$ model needs to accurately distinguish between real and generated data. Despite the challenges, GANs have shown great potential in generating high-quality and diverse samples in a range of applications, such as image synthesis, text generation, and music composition.

\subsection{{RLHF}}
{Reinforcement-learning-based training methods have shown state-of-the-art success in recent years in natural language processing tasks. One of the most advanced training methods is the reinforcement learning with hybrid feedback (RLHF) approach. This method combines the advantages of both human feedback and self-supervised learning, enabling the model to learn from its own mistakes while also benefiting from human expertise. RLHF has been successfully applied in various tasks, including machine translation, text generation, and summarization.}

{This is due to the absence of human feedback in the approach proposed in this article. So we used the feedback provided by ChatGPT~\cite{gpt}. }


\section{EvoText}
In this section, the training process of {EvoText} is introduced, and the theoretical representation and algorithmic implementation are given. 

\subsection{Priori Learning of Discriminator}
Given an a priori dataset $Data = \left \{ \left ( \boldsymbol{x}_{1},y_{1} \right ), \left ( \boldsymbol{x}_{2},y_{2} \right ), \cdots,\left ( \boldsymbol{x}_{N},y_{N} \right ) \right \} $, which could be related to tasks, such as grammar judgment or semantic rationalization, we aim to fine-tune the discriminator model using the following objective function:
\begin{eqnarray}
\min \limits_{\boldsymbol{\theta}} \mathbb{E}_{Data}\left[ \sum_{n=1}^{N}  L\left(D_{\boldsymbol{\theta}}\left(\boldsymbol{x}_{n}\right), y_{n}\right)\right]
\end{eqnarray}

Here, $D_{\theta}$ represents the pretrained natural language understanding model, and $L$ represents the loss function.

\subsection{Pre-Warm-Up Training for Generator}
We need to freeze all Transformer blocks in the pretrained natural language generative model $G$ and add new linear and softmax layers on top. We then train the newly added layers using the prior dataset through the following equation:
\begin{eqnarray}
\min \limits_{\boldsymbol{\vartheta}} \mathbb{E}_{Data}\left[ \sum_{n=1}^{N}  L\left(F_{\vartheta}(G^{blocks}_{\boldsymbol{\Theta}}\left(\boldsymbol{x}_{n})\right), y_{n}\right)\right]
\end{eqnarray}
where $G^{blocks}_{\boldsymbol{\Theta}}$ refers to all the frozen Transformer blocks of the pretrained model $G$ with parameters $\boldsymbol{\Theta}$, and $F_{\vartheta}$ represents the newly added linear layer and its trainable parameters. The objective is to minimize the expected value of the loss function $L$ on the prior dataset, where $y_n$ is the ground truth label for input example $\boldsymbol{x}_n$.

\subsection{Training Dataset}
Give a new token $\boldsymbol{Z}^{in}_{K} = \left \{  \boldsymbol{z}_{1} ,\boldsymbol{z}_{2}, \cdots,\boldsymbol{z}_{k} \right \} $ representing the input generative model $G$, such as \textit{``Yesterday''}, the token is fed into the $G$ model to obtain $\boldsymbol{Z}^{out}_{K} = \left \{  \boldsymbol{z}_{1},\boldsymbol{z}_{2}, \cdots,\boldsymbol{z}_{k+n} \right \} $, e.g., \textit{``Yesterday, a man named Jack said he saw an alien, the skies near New Orleans''.} It can express as $\boldsymbol{Z}^{out}_{K} = G_{\boldsymbol{\Theta }}(\boldsymbol{Z}^{in}_{K})$. The generated token $\boldsymbol{Z}^{out}_{K}$ is fed into the discriminator model $D$ to obtain the label $Y_{K} = D_{\theta}(\boldsymbol{Z}^{out}_{K})$ for each token. Suppose we need to construct M samples, and finally, we obtain the posterior data as $\boldsymbol{Z} = \left \{ \left ( \boldsymbol{Z}^{out}_{1}, Y_{1} \right ), \left ( \boldsymbol{Z}^{out}_{2}, Y_{2} \right ), \cdots,\left ( \boldsymbol{Z}^{out}_{M}, Y_{M} \right ) \right \}$.

\subsection{Supervised Fine-Tuning for Generators}

Supervised training has a greater impact on the model than unsupervised training, even when using the same dataset size. To fine-tune the parameters of all Transformer blocks in the generator $G^{blocks}_{\boldsymbol{\Theta }}$, we enable the gradient of all generator parameters. The optimization objective is defined as follows:

\begin{eqnarray}
\begin{split}
\min \limits_{\boldsymbol{\Phi }} \mathbb{E}_{\left ( \boldsymbol{Z}^{out},Y \right ) \sim \boldsymbol{Z} }\left[ \sum_{n=1}^{M}  L\left(F_{\vartheta }(G^{blocks}_{\boldsymbol{\Theta }}\left(\boldsymbol{Z}^{out}_{n})\right), Y_{n}\right)\right]
\end{split}
\end{eqnarray}
where $\boldsymbol{\Phi}$ denotes the parameters of the model $G^{blocks}_{\boldsymbol{\Theta }}$, and $\boldsymbol{\vartheta}$ represents the parameters of the linear layer.

\subsection{Semi-Supervised Fine-Tuning for Generators}

Assuming that grammatical sentences are labeled as $Y_{K}=1$ and ungrammatical sentences are labeled as $Y_{K}=0$, we extract all tokens with $Y_{K}=1$ in the dataset  $\boldsymbol{Z}$ as $\boldsymbol{Z}^{fit}= \left \{  \boldsymbol{z}_{1} ,\boldsymbol{z}_{2}, \cdots,\boldsymbol{z}_{m} \right \}$. The generator model $G$ is again unsupervised pretrained using corpus $\boldsymbol{Z}^{fit}$. The generator model $G$ is then unsupervised pretrained using the corpus $\boldsymbol{Z}^{fit}$, where $\boldsymbol{z{k}} = (u_{1},u_{2},\cdots,u_{n})$ represents an unsupervised token. To accomplish this, we use autoregressive language modeling to maximize the following likelihood function:
\begin{eqnarray}
L_{1}\left ( Z^{fit} \right )  = \sum_{i}^{} log G\left ( u_{i}\mid u_{i-k},\dots,u_{i-1};{\Theta } \right ) 
\end{eqnarray}
where $k$ is the size of the context window.

\subsection{{Self-Escalation}} \label{UP}
\begin{figure}[H]
\includegraphics[width=13.5 cm]{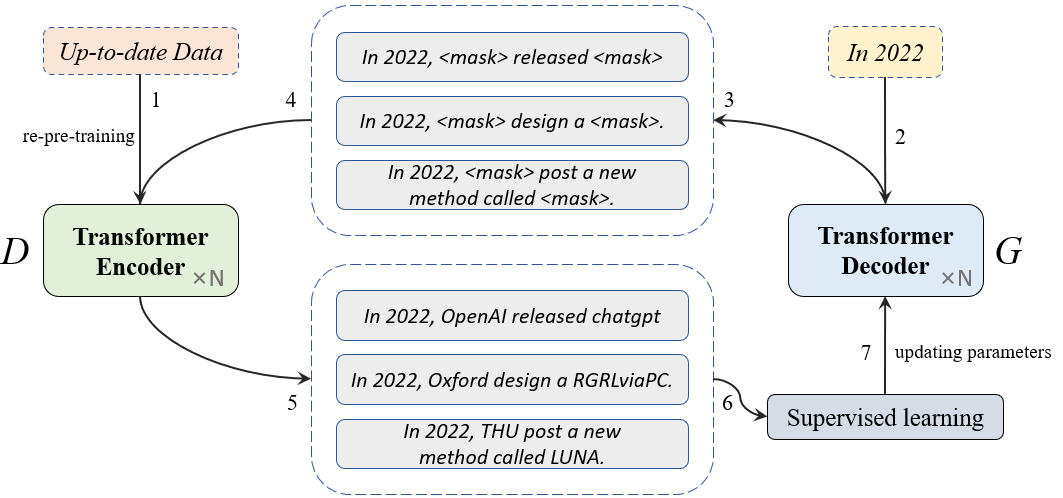}
\caption{The process of learning up-to-date knowledge. The probability of being masked in this case is 15\%, which is the same proportion as in the pretraining of BERT.\label{up-to-date2}}
\end{figure}

You can adjust the discriminator model $D$ to make it perform better, and {EvoText}  
is even superior. You can also continue to pretrain the discriminator model $D$ with new knowledge to bring it up-to-date. Certain words of text in the $Z$ dataset are randomly masked, and the masked words are predicted using the $D$ model, which is then input to $G$ for supervised fine-tuning. {As illustrated in Figure \ref{up-to-date2}, the proposed method consists of several steps. First, the up-to-date data is retrained to update the discriminator. Second, the generator is given a command containing a specific year (e.g., ``In 2022'') to generate the data. Third, 15\% of the words in the generated data is masked. Fourth, the discriminator performs word completion on the masked words. Finally, the completed data are subjected to a supervised fine-tuning of the generator, which is labeled one (by default, all statements are grammatically correct after the discriminator's completion).}

\subsection{Algorithm Implementation}

As shown in Algorithms~\ref{Priori} and \ref{training}, this is the full process of {EvoText}. 

\begin{algorithm}[H]
	\caption{Priori Learning.} 
	\label{Priori} 
	\begin{algorithmic}[1]
		\REQUIRE Training samples $Data = \left \{ \left ( \boldsymbol{x},y \right ) \right \} $, fine-tuning step $N_{ep}$, learning rate $\tau_{1}$ for $D$, and learning rate $\tau_{2}$ for $G$.
        \STATE Initialize $\theta$ and $\vartheta$
        \FOR{$t = 1\cdots N_{ep}$}
        \FOR{minibatch $B \subset Data$}
        \STATE $g_{1} \leftarrow \mathbb{E}_{B}\left[ \sum_{n=1}^{N}  \bigtriangledown_{\boldsymbol{\theta}} L\left(D_{\boldsymbol{\theta}}\left(\boldsymbol{x}_{n}\right), y_{n}\right)\right] $
        \STATE $\theta \leftarrow \theta - \tau_{1} g_{1}$
        \STATE $g_{2} \leftarrow  \mathbb{E}_{B}\left[ \sum_{n=1}^{N} \bigtriangledown_{\boldsymbol{\vartheta}}  L\left(F_{\vartheta }(G^{blocks}_{\boldsymbol{\Theta }}\left(\boldsymbol{x}_{n})\right), y_{n}\right)\right]$
        \STATE $\vartheta \leftarrow \vartheta - \tau_{2} g_{2}$
        \ENDFOR
        \ENDFOR
	\end{algorithmic} 
\end{algorithm}
\vspace{-9pt}
\begin{algorithm}[H]
	\caption{Training Process.} 
	\label{training} 
	\begin{algorithmic}[1]
		\REQUIRE Input token $\boldsymbol{Z}^{in}_{M} = \left \{\boldsymbol{Z}_{K} \right \}=(z_{k}) $, sequence maximum length $l$, termination character $\alpha$, minibatch size $M$, Supervised Fine-tuning learning rate $\tau_{3}$, Semi-supervised Fine-tuning learning rate $\tau_{4}$.
        \STATE Initialize $\Theta$ and $\Phi$.
        \FOR{$b = 1\cdots M$}
        \FOR{$t = 1\cdots l$}
        \STATE $z_{k+t} \leftarrow G_{\boldsymbol{\Theta }}(\boldsymbol{Z}_{b}) $
        \IF{$z_{k+t}=\alpha$}
        \STATE break
        \ENDIF 
        \STATE $K \leftarrow K+1 $
        \STATE $\boldsymbol{Z}_{b} \leftarrow (z_{k-t},z_{k}) $
        \ENDFOR
        \ENDFOR
        \STATE $\boldsymbol{Y} \leftarrow D_{\boldsymbol{\theta }}(\boldsymbol{Z}^{in}_{M}) $
        \STATE $\boldsymbol{Z}^{out}_{M} = \left \{(\boldsymbol{Z}^{in}_{M},\boldsymbol{Y}) \right \}$
        \IF{Supervised Fine-tuning}
        \STATE The cross-entropy loss function L.
        
        \STATE $g \leftarrow \mathbb{E}_{B}\left[ \sum_{n=1}^{M} \bigtriangledown_{\boldsymbol{\Phi}} L\left(F_{\vartheta }(G^{blocks}_{\boldsymbol{\Theta }}\left(\boldsymbol{Z}^{out}_{n})\right), Y_{n}\right)\right]$
        \STATE $\Phi \leftarrow \Phi - \tau_{3} g$
        \ENDIF
        \IF{Semi-upervised Fine-tuning}
        \STATE Unsupervised loss function L1.
        \FOR{$n=1 \dots m$}
        \IF{$Y_{n}=1$}
        \STATE $g \leftarrow \mathbb{E}_{B}\left[ \bigtriangledown_{\boldsymbol{\Theta}} L1\left(G_{\boldsymbol{\Theta }}\left(\boldsymbol{Z}^{out}_{n}\right), {Z}^{out}_{n}\right)\right]$
        \ENDIF
        \STATE $\Theta \leftarrow \Theta - \tau_{4} g$
        \ENDFOR
        \ENDIF
	\end{algorithmic}
\end{algorithm}

\section{Experimental Setup}
In this section, we describe our experimental setup to demonstrate the performance of our algorithm. 

\subsection{Experimental Environment}
We utilize a server configuration consisting of a 120-core Xeon(R) Platinum 8358P CPU @ 2.60 GHz and 8 NVIDIA A100 (80 GB) GPUs. In order to ensure optimal efficiency, we release the GPU when the model is not being trained after deploying it in a real-world~scenario.

\subsection{Experimental Model}
To demonstrate the performance of {EvoText}, we adopted the GAN approach and selected 4 natural language understanding models and 8 natural language {generation models.} 
 ({These models are part of the PyTorch-Transformers library:~\url{https://github. com/huggingface/pytorch-transformers} (accessed on 17 February 2023)}).

\subsubsection{BERT}
The BERT model is widely recognized as one of the most outstanding models in recent years, having topped the GLUE tasks list when it was first released \cite{wang-etal-2018-glue}. For this experiment, we exclusively utilize the large-cased version of BERT as the discriminator model. We apply the {EvoText} method to the fine-tuning of this model. Notably, the BERT $_{large-cased}$ 
 model boasts 16 layers of Transformer encoders, 24 self-attentive heads, and \mbox{330 million} parameters, while the BERT$_{base-cased}$ model has 12 layers of Transformer encoders, 12~self-attentive heads, and 104 million parameters.

\subsubsection{RoBERTa}
The RoBERTa model is an improved version of the BERT model that requires longer training time, a larger batch size, and more training data. Unlike BERT, RoBERTa uses dynamic masking and text encoding, moving away from BERT's NSP task. It modifies the key hyperparameters in BERT based on BERT's language masking strategy, resulting in better generalization to downstream tasks. Despite these modifications, the overall number of parameters in RoBERTa is consistent with BERT.

\subsubsection{GPT-2}
The primary contribution of GPT-2 is its exploration of the performance of larger-scale models in ZERO-SHOT scenarios, where no fine-tuning is used. With only pretraining, hints, and predictions, GPT-2 achieved state-of-the-art results in 8 out of 9 tasks. Additionally, it is an exceptional model for natural language generation. The GPT-2$_{small}$ model and BERT$_{medium}$ have 24 and 12 layers of Transformer decoders, 24 and 12 self-attentive heads, and 335M and 124M parameters, respectively. Moreover, GPT-2 also offers larger models, such as GPT-2$_{large}$ with 774M parameters and GPT-2$_{xl}$ with 1.5B parameters.

\subsubsection{GPT-Neo}
The GPT-Neo~\cite{gpt-neo,gao2020pile} 1.3B is a Transformer model trained on the Pile using cross-entropy loss. As an autoregressive language model, it learns to predict the next token in a given sequence of English text, thereby capturing internal representations of English. These representations can then be used to extract features that are useful for downstream tasks. Although language models have many applications beyond this, there are still many unknowns in this area of research. 

\subsubsection{OPT}
The OPT model~\cite{iyer2022opt} is primarily pretrained using a causal language model (CLM) target, which belongs to the family of GPT-3 models. The pretrained model can be used to evaluate prompts and generate text for downstream tasks. Additionally, the model can be fine-tuned on downstream tasks using CLM instances. The experiments in this paper use models with 125M and 350M parameters.

\subsubsection{Transformer-XL}
The Transformer-XL model~\cite{transxl} introduces two innovations to the Vanilla Transformer, a recurrence mechanism and relative positional coding. An additional advantage of Transformer-XL over Vanilla Transformer is that it can be used for word-level and character-level language modeling. It achieves state-of-the-art language modeling results on several different datasets and combines a circularity mechanism with an attention mechanism that allows the model to learn long-term dependencies.

\subsubsection{{Language Models with Pre-trained Word Embeddings and without Pre-Trained Word~Embeddings}}
{In addition to attention-based models, pretrained word embedding models such as Word2Vec~\cite{Word2Vec} or Glove~\cite{Glove} can also yield good results when incorporated into the word embedding layer. Similarly, scratch-trained word embedding layers can be effective for specific tasks, such as hate detection or text toxicity detection~\cite{subba2022heterogeneous,rodriguez2022word,siino2022fake,saleh2021detection,siino2021detection,incitti2023beyond}. In this paper, we evaluate and compare various pretrained and scratch-trained word embedding models as discriminators to assess their impact on the overall training system.}

\subsection{Dataset}
In the subsequent experiments, we chose an a priori dataset for discriminators.

\subsubsection{CoLA}
The CoLA (corpus of linguistic acceptability) \cite{warstadtetal2019neural} consists of 10,657 sentences from 23 linguistic publications, professionally annotated for acceptability (grammaticality) by their original authors. The public version presented here contains 9594 sentences {from} the training and development sets, excluding 1063 sentences {from} the retention test set. The goal is to classify the sentences into either acceptable or unacceptable categories based on their grammaticality. Due to its carefully curated and annotated nature, CoLA is a valuable resource for evaluating the performance of various NLP models and techniques in the domain of language understanding.

\subsubsection{LAMBADA}
The source of the corpus constructed by LAMBADA~\cite{papernEtAl2016P161} is unpublished anthologies. The rationale is to minimize the influence of generic knowledge on the answers, i.e., it is difficult for the model to derive answers from generic knowledge. It consists of 5325 novels and 465 million words. {LAMBADA has been widely used for language generation tasks and language understanding tasks, such as language modeling and text comprehension, where the goal is to predict the next word in a given sentence based on the \mbox{preceding context.}}

\subsubsection{CBT}
The children's book test (CBT) aims to directly measure the extent to which language models exploit the wider language environment. The CBT is built from books that are freely accessible. {The CBT has been widely used for evaluating the performance of various NLP models and techniques in the domain of language understanding and~generation.}

\subsubsection{WikiText}
The WikiText~\cite{merity2016pointer} dataset is a large-scale language modeling dataset that is widely used in natural language processing research. It is created by extracting articles from the English Wikipedia and is available in three versions: WikiText-2, WikiText-103, and WikiText-500k. The dataset includes articles covering a wide range of topics, providing a diverse range of text for training and evaluation. The WikiText dataset has been used in various language modeling tasks, including next word prediction, text generation, and text classification. It is a valuable resource for training and evaluating natural language processing models, and its use has contributed significantly to the development of language modeling research.

\subsubsection{PTB}
The PTB(penn treebank dataset)~\cite{marcusetal1993building} contains 42,000, 3000, and 3000 English sentences for the training set, validation set, and test set. \textit{``<sos>''} is the start signal of each sentence, and \textit{``<eos>''} is the end signal of each sentence. {The dataset is annotated with part-of-speech tags, constituency parse trees, and semantic roles, providing rich linguistic annotations for various natural language processing tasks. The PTB has been used in a wide range of natural language processing tasks, including language modeling, part-of-speech tagging, named entity recognition, parsing, and machine translation.}

\subsubsection{enwiki8 and text8}
The {text8}~
({the dataset is available for download at \url{https://huggingface.co/datasets/enwik8} (accessed on 17 February 2023)}) comes from {enwiki8}~
({the dataset is available for download at \url{http://mattmahoney.NET/dc/text8.zip} (accessed on 17 February 2023)}), which was first used to conduct text compression. Simply put, enwiki8 is the first 100,000,000 characters picked up from Wikipedia; and text8 is the result of removing all kinds of strange symbols and non-English characters from these characters, then converting uppercase characters into lowercase characters and transforming numbers into the corresponding English words. {This dataset aims to learn distributed representations of words that capture their semantic and syntactic relationships, and it has been used in various natural language processing tasks, including language modeling, text generation, and word embeddings.}

\subsubsection{1BM}
The 1BW {(one billion word)} 
~\cite{chelba2013one} dataset is a large English language corpus used for pretraining language models. It contains one billion words and is freely available for research purposes. This benchmark dataset is widely used for evaluating the performance of statistical language models and is composed of various genres and topics, including news, technology, and novels. It was proposed by the Google Brain team and is considered a standard for measuring progress in the field of natural language processing. {The 1BW dataset has been used for pretraining language models to improve their performance on downstream NLP tasks, such as text classification, sentiment analysis, and \mbox{language~generation.}} 

\subsection{Model Evaluation Indicators}
We use PPL (perplexity), ACC (accuracy), and BPC (bits-per-character) as performance metrics for our experiments. PPL measures the average number of choices available to the model when predicting the next word in a sentence and is calculated using the following formula:
\begin{equation}
\text{PPL}(S) = \sqrt[m]{\frac{1}{p(w_{1},w_{2},\dots,w_{m})}} \\
 = \sqrt[m]{\prod_{i=1}^{m} \frac{1}{p(w_{i}\mid w_{1},\dots ,w_{i-1} )}}
\end{equation}
where $S$ is the sentence being evaluated, $m$ is the length of the sentence, and $p(w_{i}\mid w_{1},\dots ,w_{i-1})$ is the probability of the $i$-th word given the preceding words in the sentence. A lower PPL value indicates better model performance.

ACC measures the percentage of correct judgments out of all judgment cases and is calculated using the following formula:

\begin{equation}
\text{ACC} = \frac{\text{TP}+\text{TN}}{\text{TP}+\text{TN}+\text{FP}+\text{FN}}
\end{equation}
where TP (true positive) is the number of cases correctly judged as positive, TN (true negative) is the number of cases correctly judged as negative, FP (false positive) is the number of cases incorrectly judged as positive, and FN (false negative) is the number of cases incorrectly judged as negative.

{In the this work, accuracy refers to the percentage of correctly predicted tokens in the test dataset. In other words, it measures how often the model predicted the correct next word given the previous words in the sentence. This metric is commonly used to evaluate the performance of language models.}

BPC measures the number of bits required on average to encode each character in the text and is calculated using the following formula:

\begin{equation}
\text{BPC} = -\frac{1}{m}\sum_{i=1}^{m} \log_{2}{p(w_{i}\mid w_{1},\dots ,w_{i-1})}
\end{equation}
where $m$ is the length of the text, and $\log_{2}{p(w_{i}\mid w_{1},\dots ,w_{i-1})}$ is the number of bits required to encode the $i$-th character given the preceding characters in the text. A lower BPC value indicates better model performance.

{Specifically, BPC measures the number of bits needed to encode each character in the text. Lower BPC values indicate better compression, which in turn indicates that the model has learned to better capture the patterns and structure of the text.}

\section{Experimental Procedure}
This section shows the parameters that need to be tuned in the actual training and the comparison with other models. 

\subsection{Data Preprocessing}

In this paper, we preprocessed the data using common techniques such as regular expression substitution and expanding English abbreviations. Table \ref{PreProcess} shows the details of the preprocessing steps.

\begin{CJK*}{UTF8}{gbsn}
\begin{table}[H]
\caption{\label{PreProcess}
Example of data preprocessing for the training set in downstream tasks. 
}

\begin{adjustwidth}{-\extralength}{0cm}
\newcolumntype{C}{>{\centering\arraybackslash}X}
\begin{tabularx}{\fulllength}{ll}
\toprule
\multicolumn{1}{c}{\textbf{Operation}}       & \multicolumn{1}{c}{\textbf{Text}}                                                                         \\ \midrule
\multirow{10}{*}{Original Text}       & our friends {\color[HTML]{329A9D} {\underline{won't}}} buy this analysis, let alone the next one we propose.   \\
                                     & one more pseudo generalization and {\color[HTML]{329A9D} {\underline{ i 'm}}} giving up . \\
                                     & one more pseudo generalization or {\color[HTML]{329A9D} {\underline{ i 'm}}} giving up .                                                                         \\
                                     & i {\color[HTML]{329A9D} {\underline{ 'll }}} fix you a drink .          \\
                                     & {\color[HTML]{329A9D} {\underline{ we 're }}} dancing the night away . \\
                                     & My little ️ ️ ️ ️ ️ \#ObsessedWith MyDog{\color[HTML]{329A9D} {\underline{ @ Cafe}}} Solstice Capitol Hill\\
                                     & More \#tinyepic things {\color[HTML]{329A9D} {\underline{  \#tinyepicwestern }}}, this one is crazy   {\color[HTML]{329A9D} {\underline{ @user }}} I may be one of your… \\
                                     & Last night ️{\color[HTML]{329A9D} {\underline{ @ Omnia }}} Night Club At Caesars Palace\\
                                     & friendship at its finest.  {\color[HTML]{329A9D} {\underline{ ....\#pixar \#toystory \#buzz \#woody \#friends \#friendship \#bff… }}}\\
                                     & I L VE working for a  {\color[HTML]{329A9D} {\underline{  cause! Yesterday's }}} balloon decor for SNN 11th Annual Back 2 School Health  {\color[HTML]{329A9D} {\underline{ … }}}\\
                              \midrule
\multirow{10}{*}{After preprocessing} & our friends {\color[HTML]{32CB00} { \underline{will not}}} buy this analysis let alone the next one we propose            \\
                                     & one more pseudo generalization and {\color[HTML]{32CB00} {\underline{ i am}}} giving up  \\
                                     & one more pseudo generalization or {\color[HTML]{32CB00} {\underline{ i am}}} giving up                                                                         \\
                                     & i {\color[HTML]{32CB00} {\underline{ will }}} fix you a drink          \\
                                     & {\color[HTML]{32CB00} {\underline{ we are }}} dancing the night away  \\
                                     & My little ObsessedWithMyDog {\color[HTML]{32CB00} { \underline{  Cafe }}} Solstice Capitol Hill \\
                                     & More tinyepic things  {\color[HTML]{32CB00} { \underline{  tinyepicwestern }}} this one is crazy  {\color[HTML]{32CB00} { \underline{  user }}} I may be one of your\\
                                     & Last night {\color[HTML]{32CB00} { \underline{  Omnia }}} Night Club At Caesars Palace\\
                                     & friendship at its finest {\color[HTML]{32CB00} { \underline{  pixar toystory buzz woody friends friendship bff }}}\\
                                     & I L VE working for a {\color[HTML]{32CB00} { \underline{  cause Yesterdays }}} balloon decor for SNN 11th Annual Back 2 School Health\\
                                     \bottomrule
\end{tabularx}\end{adjustwidth}
\end{table}

\end{CJK*}

\subsection{Fine-Tuning of Discriminators For Priori Datasets}
In this paper, we employed BERT{${large}$}, BERT{${base}$}, RoBERTa{${large}$}, or RoBERTa{${base}$} as the discriminator model. Since these models are pretrained, they need to be fine-tuned to achieve optimal results in downstream tasks. During the fine-tuning process, it is recommended to use a lower learning rate and a smaller number of epochs to update the model. This is because using large learning rates and epochs may cause the model to fail to converge or overfit, which can negatively impact the model's performance in this task.

\begin{figure}[H]
\includegraphics[width=10.5 cm]{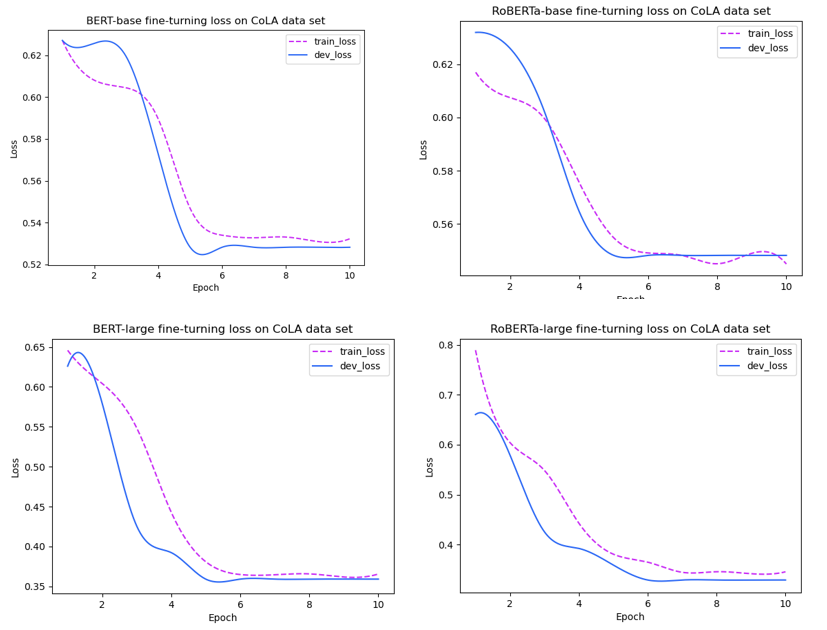}
\caption{BERT$_{large}$, BERT$_{base}$, RoBERTa$_{large}$, and RoBERTa$_{base}$ model fine-tuned loss rate in CoLA~dataset.\label{LOSS}}
\end{figure}

{{\textbf{Results}}} As illustrated in Figure~\ref{LOSS}, we observed that the large model performed better on the CoLA task, with RoBERTa exhibiting the lowest loss rate. During the pretraining process, we set the maximum length of the tokenizer to 45, enabled padding, and used a minibatch size of 512. We fine-tuned the model using several of the most commonly used parameter settings. As summarized in Table~\ref{parameter}, we achieved the best results with a learning rate of {$3 \times 10^{-5}$} 
 and 10 epochs. Following the fine-tuning process, the RoBERTa$_{large}$ model demonstrated the ability to make judgments about grammatical plausibility.

\begin{table}[H]
\caption{Setting different parameters allows the RoBERTa$_{large}$ model to fine-tune the loss rate of the validation set in the CoLA task.\label{parameter}}
\newcolumntype{C}{>{\centering\arraybackslash}X}
\begin{tabularx}{\textwidth}{CCCCC}
\toprule
\multicolumn{2}{c}{\multirow{2}{*}{}} & \multicolumn{3}{c}{\textbf{Learning Rate}}\\\cmidrule{3-5}
\multicolumn{2}{c}{}                  & {\boldmath{$2 \times 10^{-5}$}}     & {\boldmath{$3 \times 10^{-5}$}}              & {\boldmath{$4 \times 10^{-5}$}}     \\ \midrule
\multirow{3}{*}{epoch}         & 5        & 0.378204 & 0.392302         & 0.425639  \\
                           & 10        & 0.360829 & {\textbf{0.359063}} 
 & 0.370926 \\
                           & 15        & 0.363742 & 0.362149          & 0.360642 \\ \bottomrule
\end{tabularx}
\end{table}

\subsection{Prewarm-Up Training of Generator}
During pretraining of GPT-2$_{medium}$, we followed the same data preprocessing steps as before, with one exception: GPT's tokenizer does not auto-pad sentences to the maximum length. Therefore, we used the special token \textit{``<|endoftext|>''} to pad sentences that were not long enough. Unlike the input to the BERT$_{large}$ model, the generator model used in GPT-2$_{medium}$ is an autoregressive language model that requires mask attention to model the data. As such, we needed to provide masks in the input to the GPT-2$_{medium}$ model to ensure optimal performance.

As shown in Table~\ref{ReWarm}, the best performance was achieved with a learning rate of {$1 \times 10^{-2}$} and 10 epochs. During training, we only updated the parameters of the last linear layer, which allowed the model to be easily fine-tuned for supervised tasks and prevented extensive updates to the linear layer parameters during fine-tuning.

\begin{table}[H]
\tablesize{\scriptsize}
\caption{\label{ReWarm}
With all Transformer block parameters of GPT-2 frozen, only the linear layer is trained to validate the results of the set on the CoLA task. 
}
\begin{adjustwidth}{-\extralength}{0cm}
		\newcolumntype{C}{>{\centering\arraybackslash}X}
		\begin{tabularx}{\fulllength}{CCCCCCCCCC}
			\toprule
\multirow{4}{*}{\textbf{Model Input}}             & \multicolumn{9}{c}{\textbf{Loss ↓}}                                                                             \\ \cmidrule{2-10} 
                                         & \textbf{lr = {\boldmath{$5 \times 10^{-3}$}}} & \textbf{lr = {\boldmath{$5 \times 10^{-3}$}}}  & \textbf{lr = {\boldmath{$5 \times 10^{-3}$}}}  & \textbf{lr = {\boldmath{$1 \times 10^{-2}$}}} & \textbf{lr = {\boldmath{$1 \times 10^{-2}$}}}         & \textbf{lr = {\boldmath{$1 \times 10^{-2}$}}}  & \textbf{lr = {\boldmath{$2 \times 10^{-2}$}}} & \textbf{lr = {\boldmath{$2 \times 10^{-2}$}}}  & \textbf{lr = {\boldmath{$2 \times 10^{-2}$}}}  \\ \cmidrule{2-10} 
                                         & \textbf{Epoch = 5} & \textbf{Epoch = 10} & \textbf{Epoch = 1}5 & \textbf{Epoch = 5} & \textbf{Epoch = 1}0        & \textbf{Epoch = 15} & \textbf{Epoch = 5} & \textbf{Epoch = 10} & \textbf{Epoch = 15} \\ \midrule
\multicolumn{1}{l}{Token}                & 2.3275  & 2.3275   & 2.3275   & 2.3275  & 2.3275          & 2.3275   & 2.3275  & 2.3275   & 2.3275   \\
\multicolumn{1}{l}{Token+mask attention} & 0.6910  & 0.6825   & 0.6722   & 0.6810  & \textbf{0.6721} & 0.6730   & 0.6754  & 0.6772   & 0.6772   \\ 
			\bottomrule
		\end{tabularx}
	\end{adjustwidth}
\end{table}

\subsection{Training Process}
\begin{figure}[H]
\includegraphics[width=5.5cm]{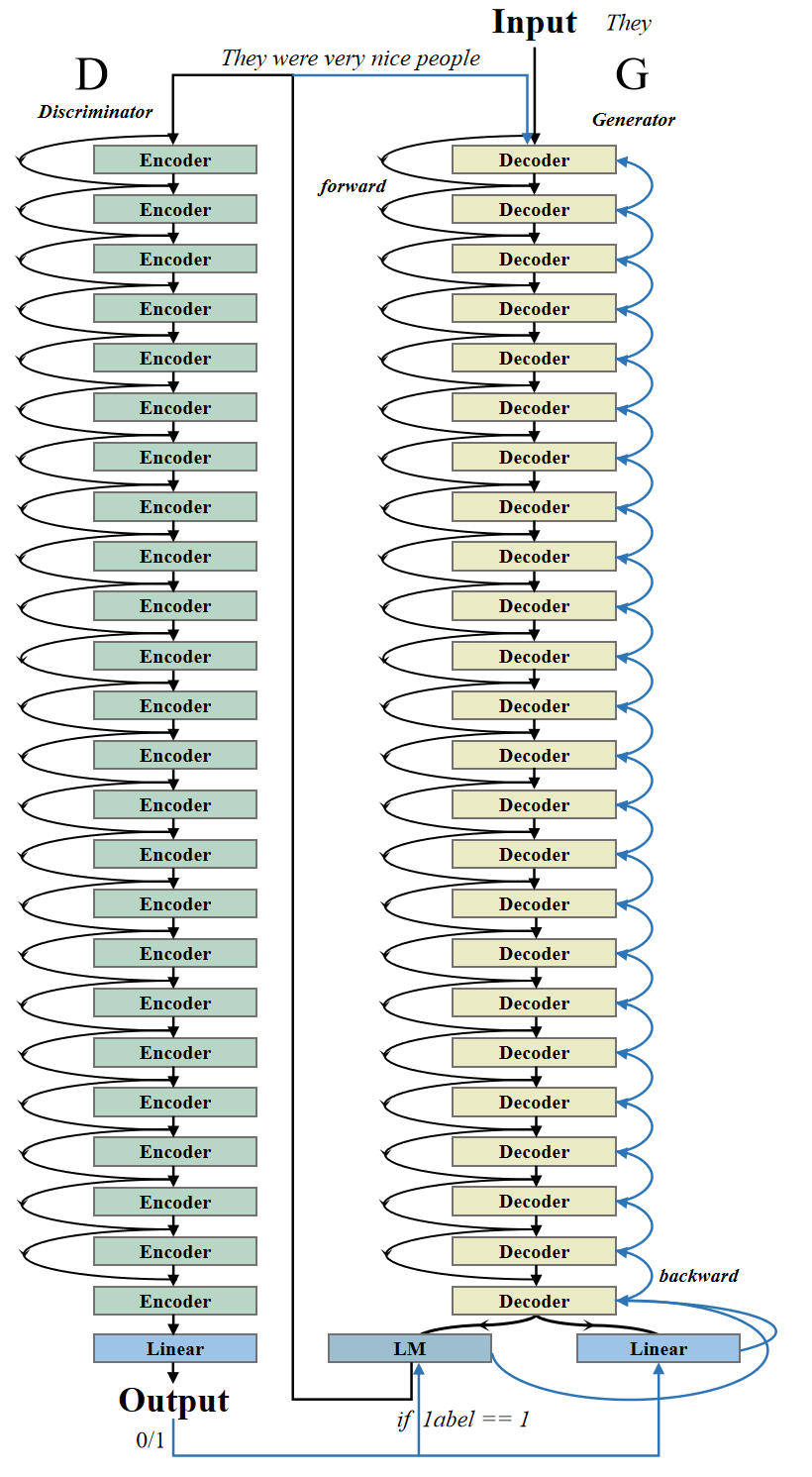}
\caption{{Training process of discriminator} 
 (BERT$_{large}$) and generator (GPT-2$_{medium}$) in EvoText. The black line indicates the forward propagation process, and the blue line indicates the reverse fine-tuning process after acquiring the label.\label{training Process}}
\end{figure}

{As illustrated in Figure}~\ref{training Process}, our framework involves training the discriminator model BERT$_{large}$ and the generator model GPT-2$_{medium}$ in a training loop. Firstly, we identify common text word-initial words and use GPT-2$_{medium}$ to complete the sentences. Subsequently, the completed sentences are fed into the fine-tuned BERT$_{large}$ model to evaluate their grammatical plausibility. This evaluation result is then utilized to conduct supervised fine-tuning of the GPT-2$_{medium}$ model. To avoid the discriminator model's errors significantly affecting the generator model, we adopt a minimal learning rate and train the generator model only one round. 

\begin{figure}[H]
\includegraphics[width=7.5 cm]{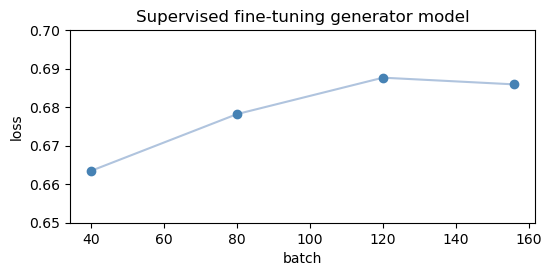}
\caption{Supervised fine-tuning of the training loss of the GPT-2$_{medium}$ model.\label{fine-turnGPT}}
\end{figure}

Table~\ref{Init data} presents some examples of text generated by GPT-2$_{medium}$. Subsequently, we fed this data to a discriminator model to assess their syntactic plausibility. The discriminator model's output value of 1 or 0 indicates whether the modified sentence is grammatically valid or invalid, respectively. Next, we used these labeled data to perform supervised fine-tuning of the GPT-2$_{medium}$ model. Figure~\ref{fine-turnGPT} shows that we used a learning rate of {$1 \times 10^{-4}$} and a minibatch size of 64 for fine-tuning.
\begin{CJK*}{UTF8}{gbsn}
\begin{table}[H]
\caption{We did not fine-tune 
the text generated by the previous GPT-2$_{medium}$ model. Then, it is input to the discriminator for judgment. The red marker denotes the data generated by the generator, while a D Output of 0 indicates that the statement is not grammatically correct, and 1 indicates that it is grammatically correct.\label{Init data}}
	\begin{adjustwidth}{-\extralength}{0cm}
		\newcolumntype{C}{>{\centering\arraybackslash}X}
		\begin{tabularx}{\fulllength}{llC}
			\toprule
\multicolumn{1}{c}{\boldmath{$G$} \textbf{Input}} & \multicolumn{1}{c}{{\color[HTML]{000000} \textbf{\boldmath{$G$} Output}}}                                         & \multicolumn{1}{c}{\textbf{\boldmath{$D$} Ouput}} \\ \midrule
That                                   & That \color[HTML]{800000} { doesn'ot have any significance, right?}                                                            & 0                                       \\
It                                     & It \color[HTML]{800000} { was a beautiful night of sunshine with some gorgeous light falling.}                                & 1                                       \\
He                                     & He \color[HTML]{800000} { said: ""{[}W{]}ith this being an attack of our religion I do feel the time will not have arrived.""} & 0                                       \\
We                                     & We \color[HTML]{800000} { have already started the implementation phase and will keep the project in mind throughout.}         & 1                                       \\
I                                      & I \color[HTML]{800000} {'ve done all those jobs.}                                                                              & 1                                       \\ \bottomrule
\end{tabularx}	\end{adjustwidth}
\end{table}
\end{CJK*}
\vspace{-12pt}

\subsection{Semi-Supervised Fine-Tuning Generator Model}
Based on the discriminator output presented in Table~\ref{Init data}, the sentences labeled with a value of 1 are fed back into the generator model for further pretraining. This process aims to enhance the generator's ability to produce grammatically correct text.

Table~\ref{Result_D} illustrates the evaluation of 10 tasks using 4 different natural language understanding models. The results indicate that RoBERTa{$_{large}$} outperforms the other models. Thus, we selected RoBERTa{$_{large}$} as the discriminator for subsequent experiments.{ As demonstrated in Table~\ref{Result_D_em}, utilizing pretrained word embeddings as a replacement for the model's word embedding layer or reinitializing the parameters to train from scratch resulted in inferior performance compared to the original RoBERTa$_{large}$ model.}

\begin{table}[H]
\tablesize{\scriptsize}
\caption{\label{Result_D}After 156 minibatch-sized EvoText sessions, we evaluated the performance of various natural language understanding models (D) and the same language generation model (GPT$_{medium}$) on 7 natural language processing tasks ZERO-SHOT. Each model was tested five times on each task, and the results were averaged.}
	\begin{adjustwidth}{-\extralength}{0cm}
		\newcolumntype{C}{>{\centering\arraybackslash}X}
		\begin{tabularx}{\fulllength}{m{2.5cm}<{\raggedright}cccccccccc}
			\toprule
\multicolumn{1}{c}{\multirow{2.5}{*}{\textbf{Model}}} & \multicolumn{2}{c}{\textbf{LAMBADA}}     & \multicolumn{2}{c}{\textbf{CBT}}                        & \multicolumn{2}{c}{\textbf{WikiText}}          & \textbf{PTB}            & \textbf{enwiki8}       & \textbf{text8}         & \textbf{1BW}            \\ \cmidrule{2-11}
\multicolumn{1}{c}{}                                & \textbf{PPL ↓}          & \textbf{ACC ↑}          & \textbf{ACC ↑ (CN)}                     & \textbf{ACC ↑ (NE)}      & \textbf{PPL ↓ (WikiText2)} & \textbf{PPL ↓ (WikiText103)} & \textbf{PPL ↓}          & \textbf{BPC ↓}         &\textbf{ BPC ↓}         & \textbf{PPL ↓}          \\ \midrule
{\textbf{Baseline}} 
                                   & \textbf{}      & \textbf{}      & \textbf{}                     &                & \textbf{}        & \textbf{}          & \textbf{}      & \textbf{}     & \textbf{}     & \textbf{}      \\
GPT-2$_{medium}$                                    & 15.60          & 55.48          & 92.35                         & 87.10          & 22.76            & 26.37              & 47.33          & 1.01          & 1.06          & 55.72          \\ \midrule
{\textbf{Ours}}                                       &                &                &                               & \textbf{}      &                  &                    &                &               &               &                \\
GPT-2$_{medium}$(G)+                                & \textbf{}      & \textbf{}      & \multicolumn{1}{l}{\textbf{}} &                & \textbf{}        & \textbf{}          & \textbf{}      & \textbf{}     & \textbf{}     & \textbf{}      \\
RoBERTa$_{large}$(D)                                & {\textbf{14.21}} & {\textbf{57.20}} & {\textbf{92.85}}                & 87.00          & {\textbf{21.94}}   & {\textbf{25.12}}     & {\textbf{45.93}} & {\textbf{1.00}} & {\textbf{1.06}} & {\textbf{55.01}} \\
RoBERTa$_{base}$(D)                                 & 14.91          & 56.21          & 92.84                         & 87.15          & 22.40            & 25.49              & 45.99          & 1.01          & 1.07          & 55.82          \\
BERT$_{large}$(D)                                   & 14.28          & 57.14          & 92.83                         & {\textbf{87.32}} & 21.97            & {\textbf{25.12}}     & 45.94          & 1.00          & 1.06          & 55.03          \\
BERT$_{base}$(D)                                    & 14.92          & 56.17          & 92.81                         & 87.02          & 21.44            & 25.32              & 45.98          & 1.01          & 1.07          & 55.80          \\			
\bottomrule
		\end{tabularx}
	\end{adjustwidth}
\end{table}

\vspace{-12pt}
\begin{table}[H]
\centering
\caption{\label{Result_D_em}After 156 minibatch-sized EvoText sessions, we evaluated the performance of various natural language understanding models (D) 
with or without pretrained word 
embeddings and the same language generation model (GPT$_{medium}$) on 7 natural language processing tasks using 
ZERO-SHOT. Each model was tested five times on each task, and the results were averaged.}
	\begin{adjustwidth}{-\extralength}{0cm}
		\newcolumntype{C}{>{\centering\arraybackslash}X}
		\begin{tabularx}{\fulllength}{m{2.5cm}<{\raggedright}cccccccccc}
			\toprule
\multicolumn{1}{c}{\multirow{2.5}{*}{\textbf{Model}}} & \multicolumn{2}{c}{\textbf{LAMBADA}}     & \multicolumn{2}{c}{\textbf{CBT}}                        & \multicolumn{2}{c}{\textbf{WikiText}}          & \textbf{PTB}            & \textbf{enwiki8}       & \textbf{text8}         & \textbf{1BW}            \\ \cmidrule{2-11}
\multicolumn{1}{c}{}                                & \textbf{PPL ↓}          & \textbf{ACC ↑}          & \textbf{ACC ↑ (CN)}                     & \textbf{ACC ↑ (NE)}      & \textbf{PPL ↓ (WikiText2)} & \textbf{PPL ↓ (WikiText103)} & \textbf{PPL ↓}          & \textbf{BPC ↓}         &\textbf{ BPC ↓}         & \textbf{PPL ↓}          \\ \midrule
\textbf{Baseline}                                   & \textbf{}      & \textbf{}      & \textbf{}                     &                & \textbf{}        & \textbf{}          & \textbf{}      & \textbf{}     & \textbf{}     & \textbf{}      \\
GPT-2$_{medium}$                                    & 15.60          & 55.48          & 92.35                         & 87.10          & 22.76            & 26.37              & 47.33          & 1.01          & 1.06          & 55.72          \\ \midrule
\textbf{Ours}                                       &                &                &                               & \textbf{}      &                  &                    &                &               &               &                \\
GPT-2$_{medium}$(G)+                                & \textbf{}      & \textbf{}      & \multicolumn{1}{l}{\textbf{}} &                & \textbf{}        & \textbf{}          & \textbf{}      & \textbf{}     & \textbf{}     & \textbf{}      \\
RoBERTa$_{large}$(D)                                & {\textbf{14.21}}  
& {\textbf{57.20}} & {\textbf{92.85}}                & 87.00          & 21.94   & {\textbf{25.12}}     & {\textbf{45.93}} & {\textbf{1.00}} & {\textbf{1.06}} & {\textbf{55.01}} \\ \midrule
RoBERTa$_{large}$+&&&&&&&&&&          \\
Word2Vec & 15.19& 57.13&  92.85&  87.03&  23.01&  27.72&  46.91&  1.03&   1.06&   55.61\\ \midrule
RoBERTa$_{large}$+&&&&&&&&&&          \\
Glove    & 15.29& 57.03&  92.79&  {\textbf{87.13}}&  23.11&  27.22&  46.93&  1.04&   1.06&   55.10\\ \midrule
RoBERTa$_{large}$+&&&&&&&&&&          \\
Scratch-trained   & 15.01& 56.13&  92.79&  87.03&  {\textbf{21.81}}&  26.42&  46.01&  1.04&   1.06&   55.10\\
\bottomrule
		\end{tabularx}
	\end{adjustwidth}
\end{table}

\subsection{Experimental Results}
After a training process consisting of 156 minibatch iterations, {which represents the average of the LAMBADA, CBT, WikiText, PTB, enwiki8, text8, and 1BW dataset sizes,} we evaluated the performance of six natural language generation models on various datasets, including LAMBADA, CBT, WikiText, PTB, enwiki8, text8, and 1BW.

Table~\ref{Result} shows that EvoText can steadily improve the performance of eight natural language generation models. Notably, the training process with just 156 steps can produce better results without altering the model architecture or the initial pretraining method. It is surprising to see that EvoText improves the performance of the GPT{$_{small}$} model to surpass that of the OPT$_{125M}$ model. These results indicate that EvoText can significantly enhance the performance of the model without requiring extensive modifications. {Our approach demonstrates favorable performance compared to the current state-of-the-art RLHF in terms of Chatgpt feedback on nearly every task. Based on the results presented in Table \ref{Result_T}, it can be observed that the EvoText training approach is highly effective in rectifying a significant portion of the grammatical errors produced by the model, which is an impressive outcome.}

\begin{table}[H]
\centering
\caption{\label{Result} {\textbf{ZERO-SHOT}} performance of the model on 7 natural language processing tasks after 156 minibatch-sized EvoText sessions. Each model was run five times on each task and averaged.}
	\begin{adjustwidth}{-\extralength}{0cm}
		\newcolumntype{C}{>{\centering\arraybackslash}X}
		\begin{tabularx}{\fulllength}{m{2.5cm}<{\raggedright}cccccccccc}
			\toprule
\multicolumn{1}{c}{\multirow{2.5}{*}{\textbf{Model}}} & \multicolumn{2}{c}{\textbf{LAMBADA}}     & \multicolumn{2}{c}{\textbf{CBT}}                        & \multicolumn{2}{c}{\textbf{WikiText}}          & \textbf{PTB}            & \textbf{enwiki8}       & \textbf{text8}         & \textbf{1BW}            \\ \cmidrule{2-11}
\multicolumn{1}{c}{}                                & \textbf{PPL ↓}          & \textbf{ACC ↑}          & \textbf{ACC ↑ (CN)}                     & \textbf{ACC ↑ (NE)}      & \textbf{PPL ↓ (WikiText2)} & \textbf{PPL ↓ (WikiText103)} & \textbf{PPL ↓}          & \textbf{BPC ↓}         &\textbf{ BPC ↓}         & \textbf{PPL ↓}          \\ \midrule
{\textbf{Baseline}}                                   &                &                &                &                &                  &                    &                &               &               &                \\
GPT-2$_{small}$                            & 35.13          & 45.99          & 87.65          & 83.4           & 29.41            & 37.50              & 65.85          & 1.16          & 1,17          & 75.20          \\
GPT-2$_{medium}$                           & 15.60          & 55.48          & 92.35          & 87.1  & 22.76            & 26.37              & 47.33          & 1.01          & 1.06          & 55.72          \\
GPT-2$_{large}$                            & 10.87          & 60.12          & 93.45          & 88.0           & 19.93            & 22.05     & 40.31          & 0.97          & 1.02          & 44.575         \\
GPT-2$_{xl}$                               & 8.63           & 63.24          & 93.30          & 89.05 & 18.34            & 17.48              & 35.76          & 0.93          & 0.98          & 42.16          \\
GPT-Neo$_{1.3B}$                           & 7.50           & 57.2           & -              & -              & \multicolumn{2}{c}{13.10}             & -              & -             & -             & -              \\
OPT$_{125M}$                               & 32.93          & 47.2           & 88.24          & 86.81          & 28.14            & 38.23              & 60.15          & 1.17          & 1,16          & 72.89          \\
OPT$_{350M}$                               & 13.29          & 56.99          & 92.25          & 87.31          & 20.99            & 25.01              & {\textbf{46.23}}          & 1.00          & 1.06          & 50.72          \\
Transformer-XL                             & -              & -              & -              & -              & -                & -                  & 54.5           & 0.99          & 1.08          & -              \\ \midrule
{\textbf{RLHF by Chatgpt}}                                   &                &                &                &                &                  &                    &                &               &               &                \\
GPT-2$_{small}$                            & 30.43          & 45.79          & 88.36          & 84.44           & 29.20            & 37.48              & 64.91          & 1.16          & 1,16          & 74.99          \\
GPT-2$_{medium}$                           & 15.29          & 55.01          & 92.41          & {\textbf{87.80}}  & 22.66            & 26.28              & 47.30          & 1.01          & 1.06          & 55.69          \\
GPT-2$_{large}$                            & 10.88          & 60.13          & 93.39          & 88.09           & {\textbf{19.23}}            & {\textbf{22.05}}     & 39.91          & 0.97          & 1.02          & 44.50         \\
GPT-2$_{xl}$                               & 8.57           & {\textbf{63.24}}          & 93.87          & 89.07 & 18.21            & {\textbf{17.47}}              & 35.75          & 0.93          & 0.98          & 42.11          \\
GPT-Neo$_{1.3B}$                           & 7.52           & 57.18           & -              & -              &15.37 & 13.00           & -              & -             & -             & -              \\
OPT$_{125M}$                               & 30.71          & 47.29           & 88.17          & {\textbf{86.99}}          & 28.12            & 37.03              & 60.24          & 1.16          & 1,16          & 70.99          \\
OPT$_{350M}$                               & 13.01          & 57.09          & 92.55          & 87.43          & {\textbf{20.81}}            & 25.01              &46.28          & 1.00          & 1.06          & 50.54          \\
Transformer-XL                             & -              & -              & -              & -              & -                & -                  & 54.25           & 0.99          & 1.08          & -              \\ \midrule
{\textbf{Ours(D is RoBERTa$_{large}$)}}    &                &                &                &                &                  &                    &                &               &               &                \\
EvoText GPT-2$_{small}$              & {\textbf{28.42}} & {\textbf{46.29}} & {\textbf{89.23}} & {\textbf{84.48}} & {\textbf{28.31}}   & {\textbf{35.72}}     & {\textbf{62.48}} & {\textbf{1.14}} & {\textbf{1,16}} & {\textbf{73.82}} \\
EvoText GPT-2$_{medium}$             & {\textbf{14.21}} & {\textbf{57.20}} & {\textbf{92.85}} & 87.0           & {\textbf{21.94}}   & {\textbf{25.12}}     & {\textbf{45.93}} & {\textbf{1.00}} &  {\textbf{1.06}} & {\textbf{55.01}} \\
EvoText GPT-2$_{large}$              & {\textbf{10.57}} & {\textbf{60.14}} & {\textbf{93.80}} & {\textbf{88.32}} & 19.93   & 22.06              & \textbf{38.20} & {\textbf{0.96}} & {\textbf{1.02}} & {\textbf{44.41}} \\
EvoText GPT-2$_{xl}$                 & {\textbf{8.09}}  & 63.23 & {\textbf{93.90}} & {\textbf{89.15}}          & {\textbf{17.92}}   & 17.50              & {\textbf{34.91}} & {\textbf{0.92}} & {\textbf{0.98}} & {\textbf{42.09}} \\
EvoText GPT-Neo$_{1.3B}$             & {\textbf{7.41}}  & {\textbf{57.19}} & \textbf{-}     & \textbf{-}     & 15.29            & {\textbf{12.81}}     & \textbf{-}     & \textbf{-}    & \textbf{-}    & \textbf{-}     \\
EvoText OPT$_{125M}$                 & {\textbf{27.51}} & {\textbf{50.09}} & {\textbf{91.01}} & 85.24 & {\textbf{28.00}}   & {\textbf{35.89}}     & {\textbf{55.99}} & {\textbf{1.14}} & {\textbf{1,15}} & {\textbf{70.21}} \\
EvoText OPT$_{350M}$                 & {\textbf{12.10}} & {\textbf{57.41}} & {\textbf{92.69}} & {\textbf{88.91}} & 21.99   & {\textbf{25.00}}     & 46.33 & {\textbf{1.00}} & {\textbf{1.06}} & {\textbf{50.02}} \\
EvoText Transformer-XL               & \textbf{-}     & \textbf{-}     & \textbf{-}     & \textbf{-}     & \textbf{-}       & \textbf{-}         &{\textbf{53.21}} & {\textbf{0.99}} & {\textbf{1.07}} & \textbf{-}     \\ \bottomrule
		\end{tabularx}
	\end{adjustwidth}
\end{table}
\vspace{-12pt} 

\begin{table}[H]
\caption{\label{Result_T}We entered ``Once upon a time'' into Baseline's GPT-2$_{xl}$ and EvoText GPT-2$_{xl}$, respectively, for comparison.}
\renewcommand{\arraystretch}{1.2}
\newcolumntype{C}{>{\centering\arraybackslash}X}
\begin{tabularx}{\textwidth}{l}
\toprule
\textbf{Generation\hspace{3.5pt}by\hspace{3.5pt}Baseline\hspace{3.5pt}GPT-2$_{xl}$}                                                                                                                                                                                                                                                                                                                                                    \\
Once upon a time, girl name is Lisa.\\ Lisa is like to go on walk in park, but yesterday, she goes on walk and she lost.\\ She asks help from man which he see, but man doesn't speak English.\\ She feels very scared and doesn't know how to come back to her house.\\ Suddenly, she saw a police car and she run to them.\\ Police helped her and she come back to her house safely.                      \\ \midrule
\textbf{Generation\hspace{3.5pt}by\hspace{3.5pt}EvoText\hspace{3.5pt}GPT-2$_{xl}$}                                                                                                                                                                                                                                                                                                                                                     \\
Once upon a time, \textbf{there\hspace{3.5pt}was\hspace{3.5pt}a\hspace{3.5pt}girl\hspace{3.5pt}named\hspace{3.5pt}Lisa.}\\ Lisa \textbf{enjoyed\hspace{3.5pt}going\hspace{3.5pt}for\hspace{3.5pt}walks\hspace{3.5pt}in\hspace{3.5pt}the\hspace{3.5pt}park}, 
but yesterday, she got lost {\textbf{while}} on a walk.\\ She asked {\textbf{for}} help from {\textbf{a}} man she {\textbf{saw}}, but he {\textbf{didn't}} speak English.\\ She \textbf{felt\hspace{3.5pt}very\hspace{3.5pt}scared} and didn't know how to get back home.\\ Suddenly, she saw a police car and \textbf{ran\hspace{3.5pt}towards\hspace{3.5pt}them.}\\ {\textbf{The}} police officers {\textbf{helped}} her and she was able to return home safely. \\ \bottomrule
\end{tabularx}
\end{table}

\subsection{Up-to-Data Knowledge Update}
We collected abstracts of preprints published on arXiv from June to September 2022 as the most up-to-date knowledge dataset. We partitioned the dataset into a training dataset, validation dataset, and testing dataset with a split ratio of 7:1:2. In the conventional methodology, the generator model is directly subjected to fine-tuning. In contrast, our methodology as shown in Section \ref{UP} entails solely fine-tuning the discriminator model, followed by EvoText training. To ensure equitable experimental outcomes, we employ the identical epoch across all trials.

Table \ref{up-to} demonstrates that the up-to-date knowledge update generator model, implemented with EvoText's approach, outperforms retraining the generator model while maintaining its performance on the ZERO-SHOT task. {However, the scalability of EvoText on larger datasets and more complex natural language processing tasks is not discussed in this paper. Nevertheless, based on the results presented in Table \ref{up-to}, it can be observed that the model not only acquires new knowledge but also avoids catastrophic forgetting of the original knowledge, which is promising for future research on scalability.}

\begin{table}[H]
\centering
\caption{\label{up-to}We evaluated the performance of both models on the arXiv dataset by retraining the discriminator at a learning rate of {$1 \times 10^{-4}$} and the generator at a learning rate of {$5 \times 10^{-5}$}.}
\newcolumntype{C}{>{\centering\arraybackslash}X}
\begin{tabularx}{\textwidth}{lCC}
\toprule
\textbf{Model}          & \textbf{PPL↓ (arXiv)} & \multicolumn{1}{l}{\textbf{PPL↓ (LAMBADA) (ZERO-SHOT)}} \\ \midrule
{\textbf{Baseline}}       &                      &                                                       \\
GPT-2$_{small}$         & 37.14   (ZERO-SHOT)             & 35.13                                                 \\
OPT$_{125M}$            & 36.80   (ZERO-SHOT)             & 32.93                                                 \\ \midrule
{\textbf{Traditonal}}     &                      &                                                       \\
GPT-2$_{small}$         & 20.42                & {\textbf{28.10}}                                        \\
OPT$_{125M}$            & 20.13                & 28.01                                                 \\ \midrule
{\textbf{Ours}}           &                      &                                                       \\
EvoText GPT-2$_{small}$ & {\textbf{19.81}}       & 28.12                                        \\
EvoText OPT$_{125M}$    & {\textbf{18.94}}       & {\textbf{28.00}}                                                 \\ \bottomrule
\end{tabularx}
\end{table}

\subsection{Ablation Experiments}
To investigate the effect of each module in EvoText, we performed 
ablation experiments by removing them one by one. These modules include the fine-tuning discriminator model, the prewarm-up generator, and the supervised and semi-supervised fine-tuning generator models.

Based on Table~\ref{Ablation}, it can be observed that each module of EvoText has an impact on the final results. Removing any of these modules may cause some negative impact on the overall performance. It is worth noting that the supervised learning module is deemed necessary in our approach.
\begin{table}[H]
\centering
\caption{\label{Ablation} We evaluated the impact of removing a particular module of EvoText GPT-2$_{media}$ on 
ZERO-SHOT performance in seven tasks.}
	\begin{adjustwidth}{-\extralength}{0cm}
		\newcolumntype{C}{>{\centering\arraybackslash}X}
		\begin{tabularx}{\fulllength}{m{2.5cm}<{\raggedright}cccccccccc}
			\toprule
\multicolumn{1}{c}{\multirow{2.5}{*}{\textbf{Model}}} & \multicolumn{2}{c}{\textbf{LAMBADA}}     & \multicolumn{2}{c}{\textbf{CBT}}                        & \multicolumn{2}{c}{\textbf{WikiText}}          & \textbf{PTB}            & \textbf{enwiki8}       & \textbf{text8}         & \textbf{1BW}            \\ \cmidrule{2-11}
\multicolumn{1}{c}{}                                & \textbf{PPL ↓}          & \textbf{ACC ↑}          & \textbf{ACC ↑ (CN)}                     & \textbf{ACC ↑ (NE)}      & \textbf{PPL ↓ (WikiText2)} & \textbf{PPL ↓ (WikiText103)} & \textbf{PPL ↓}          & \textbf{BPC ↓}         &\textbf{ BPC ↓}         & \textbf{PPL ↓}          \\ \midrule
{\textbf{Init}}                                       &                &                &                \\
Full module EvoText                           & {\textbf{14.21}} & {\textbf{57.20}} & {\textbf{92.85}} & {\textbf{87.0}}           & {\textbf{21.94}}   & {\textbf{25.12}}     & {\textbf{45.93}} & {\textbf{1.00}} & {\textbf{1.06}} & {\textbf{55.01}} \\ \midrule
\textbf{Ablation\hspace{3pt}Experiments}                       &                &                &                \\
Remove D pretraining                               & 14.80          & 55.94   &92.95 & 86.53      & 22.46  &27.54 &46.99 &1.01 &1.06 &55.72        \\
Remove G Pre-warm-up Training                       & 14.34          & 57.03  &92.90 & 86.93        & 22.12 &25.54 &45.99 &1.00 &1.06 &55.11         \\
Remove supervised fine-tuning                       & 15.12          & 55.62   &92.89 & 87.23       & 22.70  &27.64 &47.90 &1.01 &1.06 &56.11         \\
Remove semi-supervised fine-tuning                  & 14.42          & 57.04   &92.90 & 86.23       & 22.08  &25.69 &46.70 &1.00 &1.06 &55.88        \\ \bottomrule
		\end{tabularx}
	\end{adjustwidth}
\end{table}

\section{Conclusions and Future Work}
In this article, we introduced EvoText, a training process for two pretrained models aimed at addressing the challenges of insufficient sample data and computational resources, allowing models to continue learning after deployment. Through fine-tuning discriminators and prewarm-up training generators, we achieved better model performance with just 156 training steps, significantly improving performance without requiring additional training data. This approach steadily improves the performance of natural language understanding and generation tasks without changing the model structure, with the potential for even greater performance improvements over time. EvoText is an effective and scalable training model that holds great promise for low-resource NLP tasks. Our extensive experiments demonstrate the potential for improving pretrained model performance and highlight the importance of supervised learning. 

{Future research directions may include exploring the potential for EvoText in other NLP tasks and applications, investigating the impact of different discriminator and generator architectures on model performance, and further exploring the potential for continued learning after deployment in other settings. Additionally, our study highlights the importance of supervised learning in NLP and suggests that future research should continue to focus on developing effective training processes for pretrained models in low-resource~settings.}


\authorcontributions{Conceptualization, methodology, formal analysis, and software, Z.Y.; verification, Z.Y., C.Z. and H.X.; writing—original draft preparation, Z.Y., C.Z. and H.X.; writing—review and editing, Y.L.; visualization, C.Z. and H.X.; supervision, Y.L.; funding acquisition, Y.L. All authors have read and agreed to the published version of the manuscript.}

\funding{The authors gratefully acknowledge the support of the AIMTEEL 202201 Open Fund for Intelligent Mining Technology and Equipment Engineering Laboratory in Anhui Province and the Anhui Provincial Department of Education Scientific Research Key Project (Grant No. 2022AH050995). The financial assistance provided by these projects was instrumental in carrying out the research presented in this paper. We would like to thank all the members of the laboratory for their valuable support and assistance. Without their help, this research would not have been possible. Finally, we would like to express our gratitude to the Anhui Polytechnic University for providing the necessary facilities and resources for this study.}

\institutionalreview{{ The study did not require ethical approval.}
}

\informedconsent{{Not applicable.}

}

\dataavailability{All data generated by the generators in this article are placed in: \url{https://github.com/DLYuanGod/Auto-learning/blob/main/Gen.csv} (access on 10 April 2023).
}
\conflictsofinterest{The authors declare no conflict of interest.} 

\begin{adjustwidth}{-\extralength}{0cm}

\reftitle{References}

\PublishersNote{}
\end{adjustwidth}
\end{document}